\documentclass[11pt]{article}
\usepackage{colacl, epsfig}
\title{Noun-phrase co-occurrence statistics for semi-automatic semantic lexicon construction}
\author{{\bf Brian Roark} \\Cognitive and Linguistic Sciences\\Box 1978\\Brown University\\Providence, RI  02912, USA\\{\tt Brian\_Roark@Brown.edu}
        \And
	{\bf Eugene Charniak} \\Computer Science \\Box 1910\\Brown University \\Providence, RI  02912, USA\\{\tt ec@cs.brown.edu}}
\begin{document}
\maketitle
\begin{abstract}
Generating semantic lexicons semi-automatically could be a
great time saver, relative to creating them by hand.  
In this paper, we present an algorithm for extracting potential 
entries for a category from an on-line corpus, based upon a small set
of exemplars.  Our algorithm finds more correct terms and fewer
incorrect ones than previous work in this area.  Additionally, 
the entries that are generated potentially provide broader
coverage of the category than would occur to an individual
coding them by hand.  Our algorithm finds many terms not included
within Wordnet (many more than previous algorithms), and could be
viewed as an ``enhancer'' of existing broad-coverage resources.
\end{abstract}
\bibliographystyle{acl}

\section{Introduction}
Semantic lexicons play an important role in many natural language
processing tasks.  Effective lexicons must often include many
domain-specific terms, so that available broad coverage resources, such
as Wordnet \cite{Miller90}, are inadequate.  For example,
both {\it Escort\/} and {\it Chinook\/} are (among other things)
types of vehicles (a car and a helicopter, respectively), but
neither are cited as so in Wordnet.  Manually building domain-specific
lexicons can be a costly, time-consuming affair.  Utilizing existing
resources, such as on-line corpora, to aid in this task could improve
performance both by
decreasing the time to construct the lexicon and by improving its
quality.

Extracting semantic information from word co-occurrence statistics has
been effective, particularly for sense disambiguation 
\cite{Schutze92,Gale92,Yarowsky95}.  In \newcite{Riloff97},
noun co-occurrence statistics were used to indicate nominal
category membership, for the purpose of aiding in the construction of
semantic lexicons.
Generically, their algorithm can be outlined as follows:
\begin{enumerate}
\item {For a given category, choose a small set of exemplars (or `seed
words')}
\item {Count co-occurrence of words and seed words within a corpus}
\item {Use a figure of merit based upon these counts to select new
seed words}
\item {Return to step 2 and iterate {\it n\/} times}
\item {Use a figure of merit to rank words for category membership and
output a ranked list}
\end{enumerate}

Our algorithm uses roughly this same generic structure, but
achieves notably superior results, by changing the specifics of:
what counts as co-occurrence; which figures of merit to use for new seed
word selection and final ranking; the method of initial seed word
selection; and how to
manage compound nouns.   In sections 2-5 we will cover each of these
topics in turn.  We will also present some experimental results from
two corpora, and discuss criteria for judging the quality of the output.

\section{Noun Co-Occurrence}
The first question that must be answered in investigating this task is why
one would expect it to work at all.  Why would one expect that 
members of the same semantic category would co-occur in
discourse?  In the word sense disambiguation task, no such claim is made:
words can serve their disambiguating purpose regardless of part-of-speech or semantic
characteristics.  In motivating their investigations, Riloff and Shepherd 
(henceforth R\&S) cited several very
specific noun constructions in which co-occurrence between nouns of the
same semantic class would be expected, including conjunctions
(cars and trucks), lists (planes, trains, and automobiles),
appositives (the plane, a twin-engined Cessna) and noun compounds
(pickup truck).  

Our algorithm focuses exclusively on these constructions.
Because the relationship between nouns in a compound is quite
different than that between nouns in the other constructions, the algorithm
consists of two separate components:
one to deal with conjunctions, lists, and appositives; and the other
to deal with noun compounds.  All compound nouns in the former
constructions are represented by the head of the compound.  We
made the simplifying assumptions that a compound noun is a string of
consecutive nouns (or, in certain cases, adjectives - see discussion below), and
that the head of the compound is the rightmost noun.

To identify conjunctions, lists, and appositives,  
we first parsed the corpus, using an efficient statistical parser \cite{Charniak98},
trained on the Penn Wall Street Journal Treebank \cite{Marcus93}.
We defined co-occurrence in these constructions using the standard
definitions of dominance and precedence.  The relation is stipulated
to be transitive, so that all head nouns in a list co-occur
with each other (e.g. in the phrase {\it planes, trains, and
automobiles} all three nouns are counted as co-occuring with each other).
Two head nouns co-occur in this algorithm if they meet the following
four conditions:
\begin{enumerate}
\item {they are both dominated by a common {\small NP} node}
\item {no dominating {\small S} or  {\small VP} nodes are dominated by
that same {\small NP} node}
\item {all head nouns that precede one, precede the other}
\item {there is a comma or conjunction that precedes one and not the
other}
\end{enumerate}

In contrast, R\&S counted
the closest noun to the left and the closest noun to the right of a
head noun as co-occuring with it.  Consider the
following sentence from the MUC-4 \shortcite{MUC92} corpus:  
{\it ``A cargo aircraft may drop bombs and a truck may be equipped with
artillery for war.''\/}   In their algorithm, both {\it cargo\/}
and {\it bombs\/} would be counted as co-occuring with {\it aircraft\/}.
In our algorithm, co-occurrence is only counted within a noun phrase,
between head nouns that are separated by a comma or conjunction.
If the sentence had read: {\it ``A cargo aircraft, fighter plane, or
combat helicopter ...''\/}, then {\it aircraft\/}, {\it plane\/}, and
{\it helicopter\/} 
would all have counted as co-occuring with each other in our algorithm.

\section{Statistics for selecting and ranking}
R\&S used the same figure of merit both for
selecting new seed words and for ranking words in the final output.
Their figure of merit was simply the ratio of the times the noun
coocurs with a noun in the seed list to the total frequency of the
noun in the corpus.  This statistic favors low frequency nouns, and
thus necessitates the inclusion of a minimum occurrence cutoff.
They stipulated that no
word occuring fewer than six times in the corpus would be
considered by the algorithm.  This cutoff has two effects:  it reduces
the noise associated with the multitude of low frequency words,
and it removes from consideration a fairly
large number of certainly valid category members.  Ideally, one would
like to reduce the noise without reducing the number of valid nouns.
Our statistics allow for the inclusion of rare occcurances.
Note that this is particularly important given our algorithm, since we
have restricted the relevant occurrences to a specific type of structure; 
even relatively common nouns may not occur in the corpus more
than a handful of times in such a context.  

The two figures of merit that we employ, one to select and one to
produce a final rank, use the following two counts for each noun:
\begin{enumerate}
\item {a noun's co-occurrences with seed words}
\item {a noun's co-occurrences with any word}
\end{enumerate}

To select new seed words, we take the ratio of count 1 to count 2 for
the noun in question.
This is similar to the figure of merit used in R\&S,
and also tends to promote low frequency nouns.
For the final ranking, we chose the log likelihood statistic
outlined in \newcite{Dunning93}, which is based upon the co-occurrence counts of all
nouns (see Dunning for details).  This statistic essentially measures
how surprising the given pattern of co-occurrence would be if the
distributions were completely random.  For instance, suppose that two
words occur forty times each, and they co-occur twenty times in a
million-word corpus.  This
would be more surprising for two completely random distributions than
if they had each occurred twice and had always co-occurred.  A simple
probability does not capture this fact.

The rationale for using two different statistics for this task is that
each is well suited for its particular role, and not particularly well
suited to the other.  We have already mentioned that the simple ratio is
ill suited to dealing with infrequent occurrences.  It is thus a poor
candidate for ranking the final output, if that list includes
words of as few as one occurrence in the corpus.
The log likelihood statistic, we found, 
is poorly suited to selecting new seed words in an iterative algorithm
of this sort, because it promotes high frequency nouns, which can then
overly influence selections in future iterations, if they are selected
as seed words.  We termed this
phenomenon {\it infection\/}, and found that it can be so strong as to kill the
further progress of a category.  For example, if we are processing
the category {\it vehicle\/} and the word {\it artillery\/} is
selected as a seed word, a whole set of weapons that co-occur with
artillery can now be selected in future iterations.  If one of those
weapons occurs frequently enough, the scores for the words that it
co-occurs with may exceed those of any vehicles, and this effect may be
strong enough that no vehicles are selected in any future
iteration.  In addition, because it promotes high frequency terms,
such a statistic
tends to have the same effect as a minimum occurrence cutoff, i.e. few
if any low frequency words get added.  A simple probability is a much more
conservative statistic, insofar as it selects far fewer words with the
potential for infection, it limits the extent of any infection that
does occur, and it includes rare words.
Our motto in using this statistic for selection is, ``First do no
harm.''  

\section{Seed word selection}
The simple ratio used to select new seed words will tend not to select higher
frequency words in the category.  The solution to this problem is
to make the initial seed word selection from among the most frequent
head nouns in the corpus.  This is a sensible approach in any case,
since it provides the broadest coverage of category occurrences, from
which to select additional likely category members.  In a task that
can suffer from sparse data, this is quite important.
We printed a list
of the most common nouns in the corpus (the top 200 to 500), and
selected category members by scanning
through this list.  Another option would be to use head nouns identified in
Wordnet, which, as a set, should include the most common members of
the category in question.  In general, however, the strength of an
algorithm of this sort is in identifying infrequent or specialized
terms. Table 1 shows the seed words that were used for some of the categories
tested.  

\begin{table*}
\begin{tabular}{|p{1.5in}|p{4.5in}|}
\hline 
{\footnotesize\bf Crimes (MUC):}&
{\footnotesize
murder(s),
crime(s),
killing(s),
trafficking,
kidnapping(s)}
\\\hline 
{\footnotesize\bf Crimes (WSJ):}&
{\footnotesize
murder(s),
crime(s),
theft(s),
fraud(s),
embezzlement}
\\\hline 
{\footnotesize\bf Vehicle:}&
{\footnotesize
plane(s),
helicopter(s),
car(s),
bus(es),
aircraft(s),
airplane(s),
vehicle(s)}
\\\hline 
{\footnotesize\bf Weapon:}&
{\footnotesize
bomb(s),
weapon(s),
rifle(s),
missile(s),
grenade(s),
machinegun(s),
dynamite}
\\\hline 
{\footnotesize\bf Machines:}&
{\footnotesize
computer(s),
machine(s),
equipment,
chip(s),
machinery}
\\\hline 

\end{tabular}
\caption{Seed Words Used}
\end{table*}

\section{Compound Nouns}
The relationship between the nouns in a compound noun is very
different from that in the other constructions we are considering.
The non-head nouns in a compound noun may or may not be legitimate
members of the category.  For instance, either {\it pickup truck\/} or
{\it pickup\/} is a legitimate vehicle, whereas {\it cargo plane\/} is
legitimate, but {\it cargo\/} is not.  For this reason, co-occurrence
within noun compounds is not considered in the iterative portions of
our algorithm.  Instead, all noun compounds with a head that is
included in our final ranked list, are evaluated for inclusion in a
second list.  

The method for evaluating whether or not to include a noun compound in
the second list is intended to exclude constructions such as {\it
government plane\/} and include constructions such as {\it fighter
plane\/}.  Simply put, the former does not correspond to a type of
vehicle in the same way that the latter does.  We made the simplifying
assumption that the higher the probability of the head given the
non-head noun, the better the construction for our purposes.  For
instance, if the noun {\it government\/} is found in a noun compound, how
likely is the head of that compound to be {\it plane\/}?  How does
this compare to the noun {\it fighter\/}?  

For this purpose, we take two counts for each noun in the compound:
\begin{enumerate}
\item {The number of times the noun occurs in a noun compound with
each of the nouns to its right in the compound}
\item {The number of times the noun occurs in a noun compound}
\end{enumerate}

For each non-head noun in the compound, we evaluate whether or not to omit it
in the output.  If all of them are omitted, or if the
resulting compound has already been output, the entry is skipped.
Each noun is evaluated as follows:  

First, the head of that
noun is determined.  To get a sense of what is meant here, consider
the following compound:  {\it nuclear-powered aircraft carrier\/}.  In
evaluating the word {\it nuclear-powered\/}, it is unclear if this
word is attached to {\it aircraft\/} or to {\it carrier\/}.  While we
know that the head of the entire compound is {\it carrier\/}, in order
to properly evaluate the word in question, we must determine which of
the words following it is its head.
This is done, in the spirit of the Dependency Model of
Lauer \shortcite{Lauer95}, by selecting the noun to its right in the
compound with the highest probability of occuring with the word
in question when occurring in a noun compound.  (In the
case that two nouns have the same probability, the rightmost noun is
chosen.)  Once the head of the word is determined, the ratio of count 1 (with the head noun
chosen) to count 2 is compared to an empirically set cutoff.  
If it falls below that cutoff, it is
omitted.  If it does not fall below the cutoff, then it is kept (provided its
head noun is not later omitted).

\section{Outline of the algorithm}
The input to the algorithm is a parsed corpus and a set of initial seed
words for the desired category.  Nouns are matched with their plurals
in the corpus, and a single representation is settled upon for both,
e.g. {\it car(s)\/}. Co-Occurrence bigrams are collected for head
nouns according to the notion of co-occurrence outlined above.  The
algorithm then proceeds as follows:
\begin{enumerate}
\item {Each noun is scored with the selecting statistic
discussed above.}
\item {The highest score of all non-seed words is determined, and all
nouns with that score are added to the seed word list.  Then return
to step one and repeat.  This iteration continues many times, in our
case fifty.}
\item {After the number of iterations in (2) are completed, 
any nouns that were not selected as seed words are discarded.
The seed word set is then returned to its original members.}
\item {Each remaining noun is given a score based upon
the log likelihood statistic discussed above.}
\item {The highest score of all non-seed words
is determined, and all nouns with that score are added to the seed
word list.  We then return to step (5) and repeat the same number of
times as the iteration in step (2)}.
\item {Two lists are output, one with head nouns, ranked by when they
were added to the seed word list in step (6), the other consisting of
noun compounds meeting the outlined criterion, ordered by when their
heads were added to the list.}
\end{enumerate}

\section{Empirical Results and Discussion}
We ran our algorithm against both the MUC-4 corpus and the Wall
Street Journal (WSJ) corpus for a variety of categories, beginning with the
categories of {\it vehicle\/} and {\it weapon\/}, both included in the
five categories
that R\&S investigated in their paper.  Other
categories that we investigated
were {\it crimes\/}, {\it people}, {\it commercial
sites\/}, {\it states\/} (as in static states of affairs), and {\it
machines\/}.  This last category was run because of the sparse data
for the category {\it weapon\/} in the Wall Street Journal. It
represents roughly the same kind of category as weapon, namely
technological artifacts.  It, in turn, produced sparse
results with the MUC-4 corpus. Tables 3 and 4 show the top results on
both the head noun and the compound
noun lists generated for the categories we tested. 

R\&S
evaluated terms for the degree to which they are related to the
category.  In contrast, we counted valid only those entries that
are clear members of the category.
Related words (e.g. {\it crash\/} for the category {\it vehicle\/})
did not count.  A valid instance was:
(1) novel (i.e. not in the original seed set);
(2) unique (i.e. not a spelling variation or pluralization of a
previously encountered entry); and 
(3) a proper class within
the category (i.e. not an individual instance or a class based upon an
incidental feature).
As an illustration of this last condition,
neither {\it Galileo Probe\/} nor {\it gray plane\/} is a valid entry,
the former because it denotes an individual and the latter because it
is a class of planes based upon an incidental feature (color).  

In the interests of generating as many valid entries as possible, we
allowed for the inclusion in noun compounds of words tagged as
adjectives or cardinality words.  In certain occasions
(e.g. {\it four-wheel drive truck\/} or {\it nuclear bomb\/}) this is necessary to avoid
losing key parts of the compound.  Most common adjectives are dropped
in our compound noun analysis, since they occur with a wide variety of
heads.  

We determined three ways to evaluate the output of the algorithm
for usefulness.  The first is the ratio of valid entries to total
entries produced.  R\&S reported a ratio of .17 valid to total entries
for both the {\it
vehicle\/} and {\it weapon\/} categories (see table 2).
On the same corpus, our
algorithm yielded a ratio of .329 valid to total entries for
the category {\it vehicle\/}, and .36 
for the category {\it weapon\/}.
This can be seen in the slope of the graphs in figure 1.
Tables 2 and 5 give the relevant data for the categories that
we investigated.  In general, the ratio of valid to total entries fell
between .2 and .4, even
in the cases that the output was relatively small.

\begin{table*}
\begin{tabular}{|p{.53in}|p{.63in}|p{.67in}|p{.67in}|p{.67in}|p{.67in}|p{.67in}|p{.67in}|}
\hline 
\multicolumn{2}{|r|}{} & 
\multicolumn{3}{c|}{\scriptsize\bf MUC-4 corpus} & 
\multicolumn{3}{c|}{\scriptsize\bf WSJ corpus} \\\hline
{\scriptsize\bf Category} & {\scriptsize\bf Algorithm}&
{\scriptsize\bf Total Terms Generated}& {\scriptsize\bf Valid
Terms Generated} & {\scriptsize\bf Valid Terms not in Wordnet} &
{\scriptsize\bf Total Terms Generated}& {\scriptsize\bf Valid 
Terms Generated} & {\scriptsize\bf Valid Terms not in Wordnet}
\\\hline
{\scriptsize\bf Vehicle} & {\scriptsize R \& C}&{\scriptsize 249} &
{\scriptsize 82} & {\scriptsize 52} & {\scriptsize 339} &
{\scriptsize 123 } & {\scriptsize 81}\\\hline
{\scriptsize\bf Vehicle} & {\scriptsize R \& S}&{\scriptsize 200} &
{\scriptsize 34} & {\scriptsize 4} & {\scriptsize NA} &
{\scriptsize NA} & {\scriptsize NA} \\\hline
{\scriptsize\bf Weapon} & {\scriptsize R \& C}&{\scriptsize 257} &
{\scriptsize 93} & {\scriptsize 54} & {\scriptsize 150} &
{\scriptsize 17} & {\scriptsize 12}\\\hline
{\scriptsize\bf Weapon} & {\scriptsize R \& S}&{\scriptsize 200} &
{\scriptsize 34} & {\scriptsize 7} & {\scriptsize NA} &
{\scriptsize NA} & {\scriptsize NA} \\\hline

\end{tabular}
\caption{\footnotesize Valid category terms found that are not in Wordnet}
\end{table*}

\begin{figure}
\epsfig{file=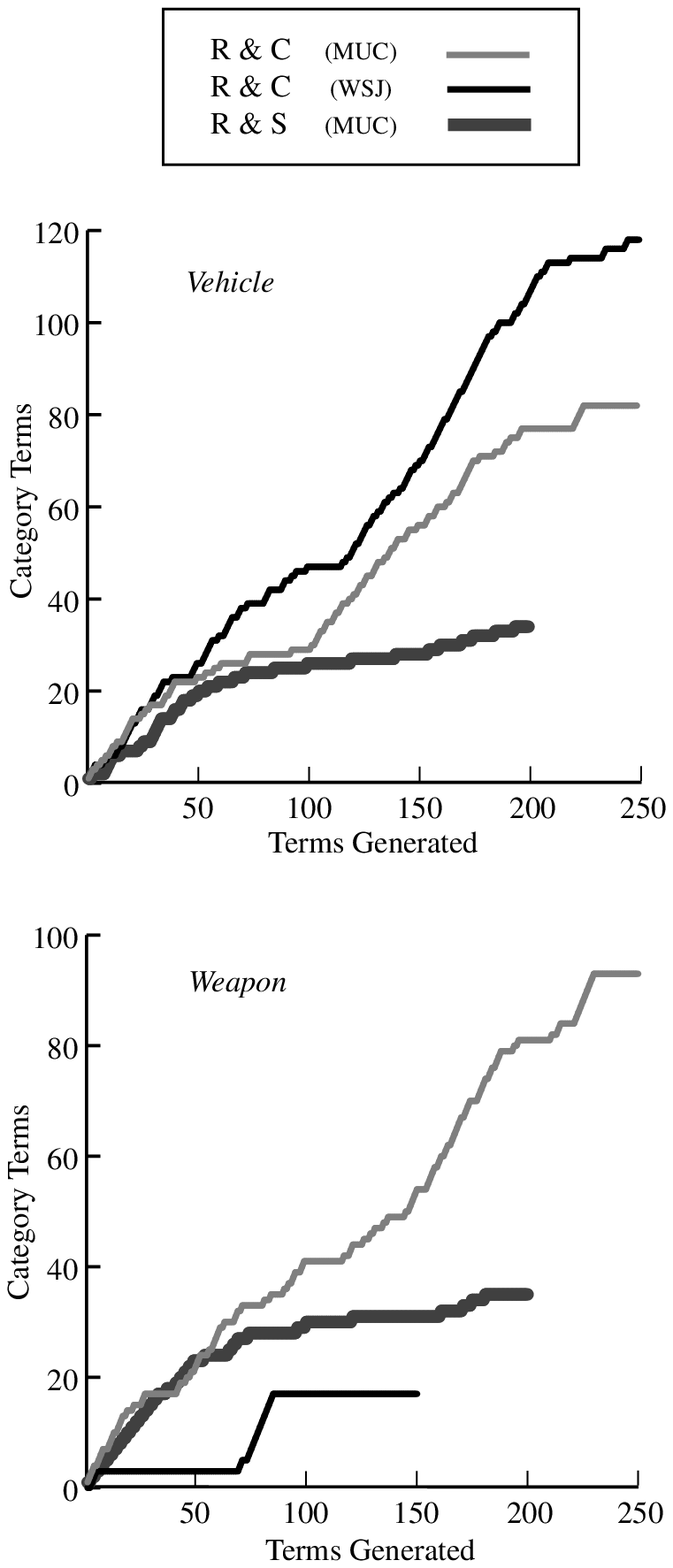, width=2.9in}
\caption{\footnotesize Results for the Categories Vehicle and Weapon}
\end{figure}

\begin{table*}
\begin{tabular}{|p{6.3in}|}
\hline 
{\footnotesize\bf Crimes (a):}
{\footnotesize
terrorism,
extortion,
robbery(es),
assassination(s),
arrest(s),
disappearance(s),
violation(s),
assault(s),
battery(es),
tortures,
raid(s),
seizure(s),
search(es),
persecution(s),
siege(s),
curfew,
capture(s),
subversion,
good(s),
humiliation,
evictions,
addiction,
demonstration(s),
outrage(s),
parade(s)}\\\hline
{\footnotesize\bf Crimes (b):}
{\footnotesize
action--the murder(s),
Justines crime(s),
drug trafficking,
body search(es),
dictator Noriega,
gun running,
witness account(s)
}\\\hline
{\footnotesize\bf Sites (a):}
{\footnotesize
office(s),
enterprise(s),
company(es),
dealership(s),
drugstore(s),
pharmacies,
supermarket(s),
terminal(s),
aqueduct(s),
shoeshops,
marinas,
theater(s),
exchange(s),
residence(s),
business(es),
employment,
farmland,
range(s),
industry(es),
commerce,
etc.,
transportation--have,
market(s),
sea,
factory(es)}\\\hline
{\footnotesize\bf Sites (b):}
{\footnotesize
grocery store(s),
hardware store(s),
appliance store(s),
book store(s),
shoe store(s),
liquor store(s),
Albatros store(s),
mortgage bank(s),
savings bank(s),
creditor bank(s),
Deutsch-Suedamerikanische bank(s),
reserve bank(s),
Democracia building(s),
apartment building(s),
hospital--the building(s)
}\\\hline
{\footnotesize\bf Vehicle (a):}
{\footnotesize
gunship(s),
truck(s),
taxi(s),
artillery,
Hughes-500,
tires,
jitneys,
tens,
Huey-500,
combat(s),
ambulance(s),
motorcycle(s),
Vides,
wagon(s),
Huancora,
individual(s),
KFIR,
M-5S,
T-33,
Mirage(s),
carrier(s),
passenger(s),
luggage,
firemen,
tank(s)}\\\hline
{\footnotesize\bf Vehicle (b):}
{\footnotesize
A-37 plane(s),
A-37 Dragonfly plane(s),
passenger plane(s),
Cessna plane(s),
twin-engined Cessna plane(s),
C-47 plane(s),
gray plane(s),
KFIR plane(s),
Avianca-HK1803 plane(s),
LATN plane(s),
Aeronica plane(s),
0-2 plane(s),
push-and-pull 0-2 plane(s),
push-and-pull plane(s),
fighter-bomber plane(s)
}\\\hline
{\footnotesize\bf Weapon (a):}
{\footnotesize
launcher(s),
submachinegun(s),
mortar(s),
explosive(s),
cartridge(s),
pistol(s),
ammunition(s),
carbine(s),
radio(s),
amount(s),
shotguns,
revolver(s),
gun(s),
materiel,
round(s),
stick(s),
clips,
caliber(s),
rocket(s),
quantity(es),
type(s),
AK-47,
backpacks,
plugs,
light(s)}\\\hline
{\footnotesize\bf Weapon (b):}
{\footnotesize
car bomb(s),
night-two bomb(s),
nuclear bomb(s),
homemade bomb(s),
incendiary bomb(s),
atomic bomb(s),
medium-sized bomb(s),
highpower bomb(s),
cluster bomb(s),
WASP cluster bomb(s),
truck bomb(s),
WASP bomb(s),
high-powered bomb(s),
20-kg bomb(s),
medium-intensity bomb(s)}
\\\hline

\end{tabular}
\caption{\footnotesize Top results from (a) the head noun list and (b) the compound
noun list using MUC-4 corpus}
\end{table*}

\begin{table*}
\begin{tabular}{|p{6.3in}|}
\hline 
{\footnotesize\bf Crimes (a):}
{\footnotesize
conspiracy(es),
perjury,
abuse(s),
influence-peddling,
sleaze,
waste(s),
forgery(es),
inefficiency(es),
racketeering,
obstruction,
bribery,
sabotage,
mail,
planner(s),
burglary(es),
robbery(es),
auto(s),
purse-snatchings,
premise(s),
fake,
sin(s),
extortion,
homicide(s),
killing(s),
statute(s)}\\\hline
{\footnotesize\bf Crimes (b):}
{\footnotesize
bribery conspiracy(es),
substance abuse(s),
dual-trading abuse(s),
monitoring abuse(s),
dessert-menu planner(s),
gun robbery(es),
chance accident(s),
carbon dioxide,
sulfur dioxide,
boiler-room scam(s),
identity scam(s),
19th-century drama(s),
fee seizure(s)}
\\\hline 
{\footnotesize\bf Machines (a):}
{\footnotesize
workstation(s),
tool(s),
robot(s),
installation(s),
dish(es),
lathes,
grinders,
subscription(s),
tractor(s),
recorder(s),
gadget(s),
bakeware,
RISC,
printer(s),
fertilizer(s),
computing,
pesticide(s),
feed,
set(s),
amplifier(s),
receiver(s),
substance(s),
tape(s),
DAT,
circumstances}\\\hline
{\footnotesize\bf Machines (b):}
{\footnotesize
hand-held computer(s),
Apple computer(s),
upstart Apple computer(s),
Apple MacIntosh computer(s),
mainframe computer(s),
Adam computer(s),
Cray computer(s),
desktop computer(s),
portable computer(s),
laptop computer(s),
MIPS computer(s),
notebook computer(s),
mainframe-class computer(s),
Compaq computer(s),
accessible computer(s)
}\\\hline
{\footnotesize\bf Sites (a):}
{\footnotesize
apartment(s),
condominium(s),
tract(s),
drugstore(s),
setting(s),
supermarket(s),
outlet(s),
cinema,
club(s),
sport(s),
lobby(es),
lounge(s),
boutique(s),
stand(s),
landmark,
bodegas,
thoroughfare,
bowling,
steak(s),
arcades,
food-production,
pizzerias,
frontier,
foreground,
mart}\\\hline
{\footnotesize\bf Sites (b):}
{\footnotesize
department store(s),
flagship store(s),
warehouse-type store(s),
chain store(s),
five-and-dime store(s),
shoe store(s),
furniture store(s),
sporting-goods store(s),
gift shop(s),
barber shop(s),
film-processing shop(s),
shoe shop(s),
butcher shop(s),
one-person shop(s),
wig shop(s)
}\\\hline
{\footnotesize\bf Vehicle (a):}
{\footnotesize
truck(s),
van(s),
minivans,
launch(es),
nightclub(s),
troop(s),
october,
tank(s),
missile(s),
ship(s),
fantasy(es),
artillery,
fondness,
convertible(s),
Escort(s),
VII,
Cherokee,
Continental(s),
Taurus,
jeep(s),
Wagoneer,
crew(s),
pickup(s),
Corsica,
Beretta}\\\hline
{\footnotesize\bf Vehicle (b):}
{\footnotesize
gun-carrying plane(s),
commuter plane(s),
fighter plane(s),
DC-10 series-10 plane(s),
high-speed plane(s),
fuel-efficient plane(s),
UH-60A Blackhawk helicopter(s),
passenger car(s),
Mercedes car(s),
American-made car(s),
battery-powered car(s),
battery-powered racing car(s),
medium-sized car(s),
side car(s),
exciting car(s)
}\\\hline

\end{tabular}
\caption{\footnotesize Top results from (a) the head noun list and (b) the compound
noun list using WSJ corpus}
\end{table*}

A second way to evaluate the algorithm is by the total number of
valid entries produced.  As can be seen from the numbers reported in
table 2, our algorithm generated from 2.4 to nearly 3
times as many valid terms for the two contrasting categories from the
MUC corpus than the algorithm of R\&S.
Even more valid terms were generated for appropriate categories using
the Wall Street Journal.

Another way to evaluate the algorithm is with the number of valid entries
produced that are not in Wordnet.  Table 2 presents these numbers for
the categories {\it vehicle\/} and {\it weapon\/}.  Whereas the R\&S
algorithm produced just 11 terms not already present in
Wordnet for the two categories combined, our algorithm produced 106,
or over 3 for every 5 valid terms produced.  It is for this reason
that we are billing our algorithm as something that could enhance
existing broad-coverage resources with domain-specific lexical information.

\section{Conclusion}
We have outlined an algorithm in this paper that, as it stands, could
significantly speed up the task of building a semantic lexicon. We
have also examined in detail the reasons why it works, and have shown
it to work well for multiple corpora and multiple categories.  The
algorithm generates many words not included in broad coverage resources,
such as Wordnet, and could be thought of as a Wordnet ``enhancer'' for
domain-specific applications.

More generally, the relative success of the algorithm demonstrates the
potential benefit of narrowing corpus input to specific kinds of
constructions, despite the danger of compounding sparse data problems.  
To this end, parsing is invaluable.

\section{Acknowledgements}
Thanks to Mark Johnson for insightful discussion and to Julie Sedivy
for helpful comments.

\begin{table}
\begin{tabular}{|p{.55in}|p{.44in}|p{.44in}|p{.44in}|p{.44in}|}
\hline {} & 
\multicolumn{2}{c|}{\scriptsize\bf MUC-4 corpus} & 
\multicolumn{2}{c|}{\scriptsize\bf WSJ corpus} \\\hline
{\scriptsize\bf Category} & 
{\scriptsize\bf Total Terms}& {\scriptsize\bf Valid
Terms} & 
{\scriptsize\bf Total Terms}& {\scriptsize\bf Valid 
Terms} 
\\\hline
{\scriptsize\bf Crimes} &{\scriptsize 115} &
{\scriptsize 24} & {\scriptsize 90} & {\scriptsize 24}\\\hline
{\scriptsize\bf Machines} &{\scriptsize 0} &
{\scriptsize 0} & {\scriptsize 335} & {\scriptsize 117}\\\hline
{\scriptsize\bf People} &{\scriptsize 338} &
{\scriptsize 85} & {\scriptsize 243} & {\scriptsize 103}\\\hline
{\scriptsize\bf Sites} &{\scriptsize 155} &
{\scriptsize 33} & {\scriptsize 140} & {\scriptsize 33}\\\hline
{\scriptsize\bf States} &{\scriptsize 90} &
{\scriptsize 35} & {\scriptsize 96} & {\scriptsize 17}\\\hline

\end{tabular}
\caption{\footnotesize Valid category terms found by our algorithm for other categories tested}
\end{table}

\bibliography{ber}
\end{document}